\documentclass{article}
\usepackage{bnaic}
\usepackage{dsfont}
\usepackage{caption}
\usepackage{subcaption}
\usepackage{graphicx}
\usepackage{hyperref}
\usepackage{url}
\usepackage{enumitem}
\usepackage{amsmath}
\usepackage{algorithmicx}
\usepackage{algpseudocode}

\newcommand{\timesteps}{500~}
\newcommand{\numepisodes}{5000~}
\newcommand{\batchsize}{10~}
\newcommand{\historylength}{10~}
\newcommand{\neurons}{100~}
\newcommand{\epochs}{2~}
\newcommand{\runs}{15~}

\setlist[itemize]{noitemsep}

\title{
    \textbf{\huge An Empirical Comparison of Neural Architectures for Reinforcement Learning in Partially Observable Environments}
}
\author{
    Denis Steckelmacher, Peter Vrancx
}
\date{\textit{
    AI-lab, Vrije Universiteit Brussel, Pleinlaan 2, 1050 Brussel
}}

\begin{document}
\ttl
\thispagestyle{empty}

\begin{abstract}
\noindent
This paper explores the performance of fitted neural Q iteration for reinforcement learning in several partially observable environments, using three recurrent neural network architectures: Long Short-Term Memory \cite{LSTM97}, Gated Recurrent Unit \cite{GRU14} and MUT1, a recurrent neural architecture evolved from a pool of several thousands candidate architectures \cite{MUT15}. A variant of fitted Q iteration, based on Advantage values \cite{AdvantageLearning96, RLLSTM01} instead of Q values, is also explored. The results show that GRU performs significantly better than LSTM and MUT1 for most of the problems considered, requiring less training episodes and less CPU time before learning a very good policy. Advantage learning also tends to produce better results.
\end{abstract}

\section{Introduction}

Reinforcement learning was originally developed for Markov Decision Processes (MDPs). It allows an agent to learn a policy to maximize a possibly delayed reward signal in a stochastic environment and guarantees convergence to an optimal policy, provided that the agent can sufficiently experiment and the environment in which it is operating is Markovian.

In many real world problems, however, the agent cannot directly perceive the full state of its environment and must make decisions based on incomplete observations of the system state. This partial observability introduces uncertainty about the true environment state and renders the problem non-Markovian from the agent's point of view. One way to deal with partially observable environments is to equip the agent with a memory of  past observations and actions in order to help it discover what the current state of the environment is. This memory can be implemented in a variety of ways, including explicit history windows \cite{i1993reinforcement,mccallum1996learning}, but this article only focuses on reinforcement learning using recurrent neural networks for function approximation. Unlike basic feed-forward networks, recurrent neural networks can contain cyclic connections between neurons. These cycles give rise to dynamic temporal behavior, which can function as an internal memory that allows these networks to model values associated with sequences of observations \cite{LSTM97, GRUvsLSTM14, RLLSTM01}. This paper aims at comparing different recurrent neural architectures when used to model value functions in a reinforcement learning context.

The next section provides necessary background on reinforcement learning and the recurrent network architectures compared in this paper. Section~\ref{sec:setup} describes the experimental setup and environments used for the comparison. The empirical results are provided in Section~\ref{sec:results}. Finally, we conclude in Section~\ref{sec:conc}. 

\section{Background}

Discrete-time reinforcement learning consists of an agent that repeatedly senses observations of its environment and performs actions. After each action $a_t \in A$, the environment changes state to $s_{t+1} \in S$ and the agent receives a reward $r_{t+1} = R(s_t, a_t, s_{t+1}) \in \mathds{R}$ and an observation $o_{t+1} \in O = f(s_{t+1})$. The agent has no knowledge of $R(s, a, s')$ and $f(s)$ and has to interact with its environment in order to learn a policy $\pi(o_t) \in \mathds{R}^{|A|}$, that gives the probability distribution of taking each of the actions for any given observation. The optimal policy $\pi^*$ is the one that, when followed by the agent, maximizes the cumulative discounted reward $r = \sum_t \gamma^t r_t$, with $0 \leq \gamma \leq 1$.

When the reward received by the agent depends solely on its current observation and action, the problem is reduced to a Markov decision process and is said to be completely observable (the agent can assume that $o_t = s_t$ without losing learning abilities). Partially observable Markov decision problems occur when the reward does not depend only on $o_t$, but on state $s_t$, whose dynamics still obey some underlying MDP, but that the agent cannot observe directly. In this case, $o_t = f(s_t)$, with $f$ an unknown one-way function part of the environment.

\subsection{Q-Learning and Advantage Learning}

Q-Learning \cite{watkins1992q} and Advantage Learning \cite{AdvantageLearning96} allow an agent to learn a policy that converges to the optimal policy given an infinite amount of time and in discrete domains.

Q-Learning estimates the $Q(o, a)$ function, that maps each state-action pair to the expected, optimal cumulative discounted reward reachable by taking action $a$ given observation $o$. At each time step, the agent observes $o_t$, takes action $a_t$ and observes $r_{t+1}$ and $o_{t+1}$. Equation \ref{eq:qlearning} is used to update the Q function after each time step, with $0 \leq \alpha \leq 1$ the learning factor.

\begin{align}
    \label{eq:qlearning}
    \delta_t          &= r_{t+1} + \gamma \max_a Q_k(o_{t+1}, a) - Q_k(o_t, a_t)\\
    Q_{k+1}(o_t, a_t) &= Q_k(o_t, a_t) + \alpha \delta_t
\end{align}

Advantage Learning \cite{AdvantageLearning96} is related to Q-Learning, but artificially decreases the value of non-optimal actions. This widens the difference between the value of the optimal action and the other ones, which allows learning to converge more easily even if the values are approximated (using function approximation). Equation \ref{eq:alearning} is used to update the Advantage values at each time step \cite{RLLSTM01}. The smaller $\kappa$ is, the widest the gap between the optimal and non-optimal actions becomes.

\begin{align}
    \label{eq:alearning}
    \delta_t          &= \max_a A_k(o_t, a) + \frac{r_{t+1} + \gamma \max_a A_k(o_{t+1}, a) - \max_a A_k(o_t, a)}{\kappa} - A_k(o_t, a_t) \\
    A_{k+1}(o_t, a_t) &= A_k(o_t, a_t) + \alpha \delta_t
\end{align}

In very large or even continuous environments, exact representation of the Q-function (or Advantage function) is no longer possible. In these cases a function approximation architecture is needed to represent the target function. It has been shown, however, that on-line Q-Learning can diverge, or converge very slowly, when used in combination with function approximation \cite{NeuralFittedQ05}. One solution to this problem is to learn the $Q$-values off-line. The method used in this paper is the neural fitted Q iteration described in \cite{NeuralFittedQ05}, an adaptation of fitted Q iteration \cite{BatchRL05} using neural networks. The agent interacts with its environment using a fixed policy until reaching the goal or a maximum number time steps have elapsed, and collects samples of the form $(o_t, a_t, r_{t+1},o_{t+1})$ . After a number of episodes have been run, the model is trained in batch on the collected data. The model maps sequences of observations to action values: $M: O^\infty \rightarrow \mathds{R}^{|A|}$.

The next subsections describe the different recurrent network architectures that we consider in this paper to represent the target functions.

\subsection{Long Short Term Memory}

An LSTM \cite{LSTM97} cell stores a value. An output gate allows the cell to modulate its output strength, while an input gate controls the intensity of the input signal that is continuously added to the cell's content. A forget gate, when set to zero, clears the content of the cell. Equations \ref{eq:lstm_gate_i} to \ref{eq:lstm_gate_o} show how the values of the gates are computed. Equations \ref{eq:lstm_cell_tilde} and \ref{eq:lstm_cell} show how to compute the value of the memory cell, and Equation \ref{eq:lstm_out} shows the output of an LSTM cell.

\begin{align}
    \label{eq:lstm_gate_i}
    i_t^j &= \sigma(W_i x_t + U_i h_{t-1})^j \\
    \label{eq:lstm_gate_f}
    f_t^j &= \sigma(W_f x_t + U_f h_{t-1})^j \\
    \label{eq:lstm_gate_o}
    o_t^j &= \sigma(W_o x_t + U_o h_{t-1})^j \\
    \label{eq:lstm_cell_tilde}
    \tilde{c}_t^j &= \tanh(W_c x_t + U_c h_{t-1})^j \\
    \label{eq:lstm_cell}
    c_t^j         &= f_t^j c_{t-1}^j + i_t^j \tilde{c}_t^j \\
    \label{eq:lstm_out}
    h_t^j &= o_t^j \tanh(c_t^j)
\end{align}

\begin{figure}
    \centering
    \includegraphics{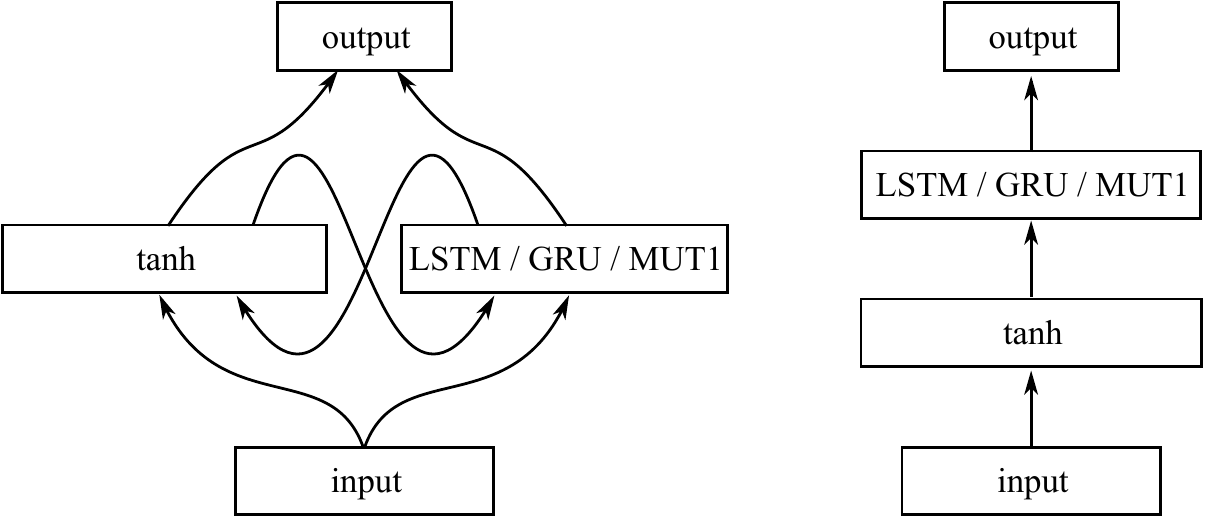}
    \caption{Cyclic architecture proposed by \cite{RLLSTM01} (left), and architecture used in this paper (right).}
    \label{fig:nnet}
\end{figure}

In \cite{RLLSTM01}, Bakker proposes a neural network architecture tailored for reinforcement learning. The network has one input neuron per observation variable, and one output neuron per action. A softmax layer transforms the output of the neural network to a probability distribution over the actions. The neural network itself consists of an LSTM layer and a simple tanh layer working in parallel: the input of the network is fed to the LSTM and tanh layer, both these layers are connected to the output, the output of the tanh layer is connected to the input of the LSTM layer and the output of the LSTM layer is connected to the input of the tanh layer (see Figure \ref{fig:nnet}). This article uses a simpler version of the network: the input is connected to a tanh layer, that is in turn connected to an LSTM layer, that is connected to the output. Both the tanh layer and the LSTM layer contain \neurons neurons (or LSTM cells). \footnote{The models themselves are built on Keras \url{http://keras.io/}, a Python library providing neural network primitives based on Theano \cite{Theano10}. Keras provides LSTM, GRU, MUT and dense fully-connected weighted layers (among others). Layers can be assembled either in a stack or in a directed acyclic graph. The connection scheme in \protect{\cite{RLLSTM01}} makes the network layer graph cyclic, and hence impossible to build using the current version of Keras.}

\subsection{Gated Recurrent Unit}

GRU has been introduced recently and follows a design completely different from LSTM \cite{GRU14, GRUvsLSTM14}. Instead of storing a value in a memory cell and updating it using input and forget gates, a GRU unit computes a candidate activation $\tilde{h}_t$ based on its input, and then produces an output that is a blend of its past output and the candidate activation. Equations \ref{eq:gru_gate_z} and \ref{eq:gru_gate_r} show how the Z (modulation) and R (reset) gates are computed. Equations \ref{eq:gru_candidate_x} and \ref{eq:gru_candidate} show how the input is mixed with the last activation in order to produce the candidate activation, and Equation \ref{eq:gru_out} shows how the last activation and the candidate activation are mixed to produce the new activation.

\begin{align}
    \label{eq:gru_gate_z}
    z_t^j &= \sigma(W_z x_t + U_z h_{t-1})^j \\
    \label{eq:gru_gate_r}
    r_t^j &= \sigma(W_r x_t + U_r h_{t-1})^j \\
    \label{eq:gru_candidate_x}
    \tilde{x}_t^j &= r_t^j h_{t-1}^j \\
    \label{eq:gru_candidate}
    \tilde{h}_t^j &= \tanh(W x_t + U \tilde{x}_t)^j \\
    \label{eq:gru_out}
    h_t^j &= (1 - z_t^j) h_{t-1}^j + z_t^j \tilde{h}_t^j
\end{align}

\subsection{MUT1}

J{\'{o}}zefowicz et al.~observed that GRU and LSTM are very different from each other, and wondered whether other recurrent neural architectures could be used. In order to discover them, they developed a genetic algorithm that evaluated thousands of recurrent neural architectures. Once the experiment was finished, they identified three architectures that performed as good as or better than LSTM and GRU on their test vectors: MUT1, MUT2 and MUT3 \cite{GRU14}.

This paper only considers MUT1, that produced the best results on preliminary experiments. Equations \ref{eq:mut_gate_z} and \ref{eq:mut_gate_r} show to compute the value of the Z and R gates, Equations \ref{eq:mut_candidate_hat} and \ref{eq:mut_candidate} show how to compute the candidate activation, and Equation \ref{eq:mut_out} shows that the output of the MUT1 uses the same type of mixing as the one used by GRU.

\begin{align}
    \label{eq:mut_gate_z}
    z_t^j &= \sigma(W_z x_t)^j \\
    \label{eq:mut_gate_r}
    r_t^j &= \sigma(W_r x_t + W_h h_{t-1})^j \\
    \label{eq:mut_candidate_hat}
    \hat{h}_t^j   &= r_t^j h_{t-1}^j \\
    \label{eq:mut_candidate}
    \tilde{h}_t^j &= \tanh(W_{\hat{h}} \hat{h}_t + \tanh(x_t))^j \\
    \label{eq:mut_out}
    h_{t} &= (1 - z_t^j) h_{t-1}^j + z_t^j \tilde{h}_t^j
\end{align}

\section{Experimental Setup}
\label{sec:setup}

In order to keep training time manageable, the neural networks are trained to associate values with the last \historylength observations, instead of the complete history. LSTM, GRU and MUT1 are able to associate values to arbitrarily long sequences of inputs, but Keras requires all the sequences on which it is trained to have the same length (possibly with padding). 

Training has to be done carefully, because one does not want the model to forget past experiences when a new batch of episodes is learned. The network has been configured to perform \epochs training epochs on the data, using a batch size of 10 (batches of 10 $O^{\historylength} \times \mathds{R}^{|A|}$ samples are used to compute an average gradient when performing backpropagation). The small number of epochs prevents the model from overfitting specific episodes.

\subsection{Environments}
\label{sec:environments}

\begin{figure}
    \centering
    \includegraphics{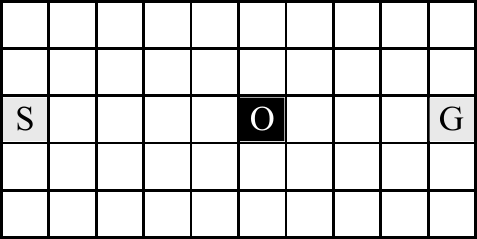}
    \caption{Grid world, the initial position (when fixed) is at S, the goal is at G, and the obstacle is depicted by O.}
    \label{fig:gridworld}
\end{figure}

Three environments are used to evaluate the neural network models. The first one is a simple fully-observable $10 \times 5$ grid world with the initial position at $(0, 2)$, the goal at $(9, 2)$ and an obstacle at $(5, 2)$ (see Figure \ref{fig:gridworld}). The agent can observe its $(x, y)$ coordinates. It receives a reward of $-1$ at each time step, $-5$ if it hits a wall or the obstacle, and $10$ when it reaches the goal.

The second environment is based on the same grid world as the first one, but the agent can only observe its $x$ coordinate. The $y$ coordinate is masked to zero.

The last environment is also based on the grid world, but the agent can only observe its orientation (whether it is facing up, down, left or right, expressed as a 0 to 3 integer number) and the distance between it and the wall in front of it. This agent-centric environment is very close to what actual robots can experience.

The ``stochastic'' variant of the experiments uses a random initial position for every episode. The agent can sense its initial $(x, y)$ coordinate at the first time step, even in otherwise partially observable environments\footnote{Some experiments have been re-run without this hint, with no change in the results. The agent learns to look left, then up, and uses those observations as initial position.}.

The observations of the agent, that consist of integer numbers, are encoded using a one-hot encoding so that they are more easily processed by neural networks. For instance, the y coordinate of the grid world can take values from 0 to 4, which are encoded as $(1, 0, 0, 0, 0), (0, 1, 0, 0, 0), ..., (0, 0, 0, 0, 1)$. For the $10 \times 5$ grid world, the neural networks therefore have 15 input neurons.

\subsection{Experiments}
\label{sec:experiments}

\begin{figure}
	\centering
	\begin{algorithmic}
		\State $\hat{Q}(s, a)$ is a neural network model
		\For{$e = 1$ to $\numepisodes$}
			\State $H_e \gets \emptyset$
			\For{$t = 1$ to $\timesteps$}
				\State Agent observes $o_t$, takes action $a_t$, receives reward $r_{t+1}$ and observation $o_{t+1}$
				\State $Q(o_t, a_t) \gets \hat{Q}(o_t, a_t) + \alpha(r_{t+1} + \gamma \max_a \hat{Q}(o_{t+1}, a) - \hat{Q}(o_t, a_t))$
				\State $H_e \gets H_e \cup \{(o_t, a_t, Q(o_t, a_t))\}$ 
			\EndFor
			\If{$e$ a multiple of \batchsize}
				\State Train $\hat{Q}(s, a)$ on $H_{e-\batchsize}, \ldots, H_e$
			\EndIf
		\EndFor
	\end{algorithmic}
	\caption{Neural fitted Q iteration as used in this paper.}
	\label{fig:neuralfittedq}
\end{figure}

Each experiment consists of \numepisodes episodes of a maximum of \timesteps time steps. During the episodes, the neural network is not trained on any new data, but $Q_{k+1}(s, a)$ values are computed based on $Q_k(s, a)$ and stored in a list. After every batch of \batchsize episodes, the neural networks are trained on the $Q_{k+1}$ values, as described in \cite{NeuralFittedQ05} and shown in Figure \ref{fig:neuralfittedq}.

The experiments themselves consist of trying to reach the goal in one of the environments described in Section \ref{sec:environments}. Each experiment is run \runs times for each combination of the following parameters:

\begin{itemize}
	\item Value iteration: Q-Learning and Advantage learning, $\alpha = 0.2$, $\gamma = 0.9$ and $\kappa = 0.3$
	\item Neural network architecture: feed-forward perceptron with a single hidden layer (nnet), LSTM (lstm), GRU (gru) and MUT1 (mut1)
	\item World: gridworld (gw), partially observable gridworld (po) and agent-centric gridworld (ac)
	\item Fixed initial position and random initial position
	\item Softmax action selection with a temperature of 0.5
\end{itemize}

\section{Empirical Results}
\label{sec:results}

Each experiment (see Section \ref{sec:experiments}) is run \runs times. The first time step at which the agent is able to maintain an average (over the 1000 next time steps) reward of more than $-15$ with a standard deviation less than 20 is called the \emph{learning time}. The best average reward obtained during a 1000-time-steps window is called the \emph{learning performance}.

\begin{table}
    \setlength{\tabcolsep}{4pt}
    \centering
    \begin{subtable}[h]{0.80\textwidth}
        \centering
        \begin{tabular}{|l|cccc|}
            \hline
            & nnet & lstm & gru & mut1 \\
            \hline
            gw & $507.3/86.6$ & $272.7/49.1$ & $\mathbf{216/37.6}$ & $4680.7/1236.8$ \\
            po & NA & $3586/1285.6$ & $\mathbf{2859.3/1495.9}$ & $4704/625.3$ \\
            ac & NA & $726.7/103.5$ & $\mathbf{664.7/96.4}$ & $1424.7/1206.2$ \\
            \hline
        \end{tabular}
        \caption{fixed initial position and Advantage Learning.}
    \end{subtable}
    \begin{subtable}[h]{0.80\textwidth}
        \centering
        \begin{tabular}{|l|cccc|}
            \hline
            & nnet & lstm & gru & mut1 \\
            \hline
            gw & $1474.7/244$ & $653.3/144.8$ & $\mathbf{540/174.5}$ & $2214.7/2055.1$ \\
            po & NA & $2844.7/1081$ & $\mathbf{1726/448.3}$ & $3808/1239.5$ \\
            ac & NA & $2829.3/1053.5$ & $\mathbf{1816/204.5}$ & $4275.3/997.7$ \\
            \hline
        \end{tabular}
        \caption{fixed initial position and Q-Learning.}
    \end{subtable}
    \begin{subtable}[h]{0.80\textwidth}
        \centering
        \begin{tabular}{|l|cccc|}
            \hline
            & nnet & lstm & gru & mut1 \\
            \hline
            gw & $\mathbf{478/211.5}$ & $514.7/286.3$ & $561.3/306.5$ & $4902.7/377$ \\
            po & NA & $4841.3/421.6$ & $\mathbf{4224/1154.9}$ & NA \\
            ac & NA & $\mathbf{818/219.4}$ & $830.7/296.9$ & $3322.7/1431.2$ \\
            \hline
        \end{tabular}
        \caption{random initial position and Advantage learning.}
    \end{subtable}
    \begin{subtable}[h]{0.80\textwidth}
        \centering
        \begin{tabular}{|l|cccc|}
            \hline
            & nnet & lstm & gru & mut1 \\
            \hline
            gw & $1021.3/210.5$ & $970.7/373.5$ & $\mathbf{664.7/141}$ & $4469.3/1400.4$ \\
            po & $4247.3/1162$ & $3146.7/1000.4$ & $\mathbf{2291.3/1236.9}$ & NA \\
            ac & NA & $\mathbf{2044/403.1}$ & $2102/889.5$ & $4504.7/787.7$ \\
            \hline
        \end{tabular}
        \caption{random initial position and Q-Learning.}
    \end{subtable}
    \caption{Learning time of all the experiments.}
    \label{fig:learningtime}
\end{table}
\begin{table}
    \setlength{\tabcolsep}{4pt}
    \centering
    \begin{subtable}[h]{0.80\textwidth}
        \centering
        \begin{tabular}{|l|cccc|}
            \hline
            & nnet & lstm & gru & mut1 \\
            \hline
            gw & $\mathbf{-0/0}$ & $-8.8/1.9$ & $-2.6/0.7$ & $-19.5/6$ \\
            po & $-26.9/2.8$ & $-15/6.1$ & $\mathbf{-9.8/2.7}$ & $-20/9.5$ \\
            ac & $-324.6/15.1$ & $-10.8/0.9$ & $\mathbf{-7.9/0.9}$ & $-10.4/0.9$ \\
            \hline
        \end{tabular}
        \caption{fixed initial position and Advantage Learning.}
    \end{subtable}
    \begin{subtable}[h]{0.80\textwidth}
        \centering
        \begin{tabular}{|l|cccc|}
            \hline
            & nnet & lstm & gru & mut1 \\
            \hline
            gw & $\mathbf{-0/0}$ & $-6.9/1.7$ & $-3.8/0.8$ & $-7.8/4.1$ \\
            po & $-24.2/1.3$ & $-12.5/2.7$ & $\mathbf{-9.2/2.3}$ & $-12.7/4$ \\
            ac & $-282.5/9.4$ & $-13.2/1.4$ & $\mathbf{-10.6/1.8}$ & $-15.6/2$ \\
            \hline
        \end{tabular}
        \caption{fixed initial position and Q-Learning.}
    \end{subtable}
    \begin{subtable}[h]{0.80\textwidth}
        \centering
        \begin{tabular}{|l|cccc|}
            \hline
            & nnet & lstm & gru & mut1 \\
            \hline
            gw & $\mathbf{3/0.1}$ & $-3.6/1.2$ & $1.8/0.3$ & $-20.1/17.9$ \\
            po & $-12.9/1.1$ & $-21.6/5.5$ & $\mathbf{-8.4/5.6}$ & $-35.1/10.8$ \\
            ac & $-187.3/15.1$ & $-5.6/1$ & $\mathbf{-3/1.8}$ & $-4.1/1.5$ \\
            \hline
        \end{tabular}
        \caption{random initial position and Advantage learning.}
    \end{subtable}
    \begin{subtable}[h]{0.80\textwidth}
        \centering
        \begin{tabular}{|l|cccc|}
            \hline
            & nnet & lstm & gru & mut1 \\
            \hline
            gw & $\mathbf{2.9/0.1}$ & $-2.8/1$ & $0.8/0.6$ & $-10.8/6.4$ \\
            po & $-9.5/0.7$ & $-8.8/2$ & $\mathbf{-3.6/1.3}$ & $-14.7/6.5$ \\
            ac & $-154.2/5.3$ & $-6.5/1.2$ & $\mathbf{-6.2/1.5}$ & $-8.6/2.7$ \\
            \hline
        \end{tabular}
        \caption{random initial position and Q-Learning.}
    \end{subtable}
    \caption{Learning performance of all the experiments.}
    \label{fig:learningperformance}
\end{table}

Table \ref{fig:learningtime} shows the learning time of the different neural networks for all the experiment configurations. Best results are emphasized in bold. Results are displayed in a mean/stddev format.

\begin{figure}
	\centering
	\begin{subfigure}[h]{0.30\textwidth}
		\centering
		\includegraphics[width=\textwidth]{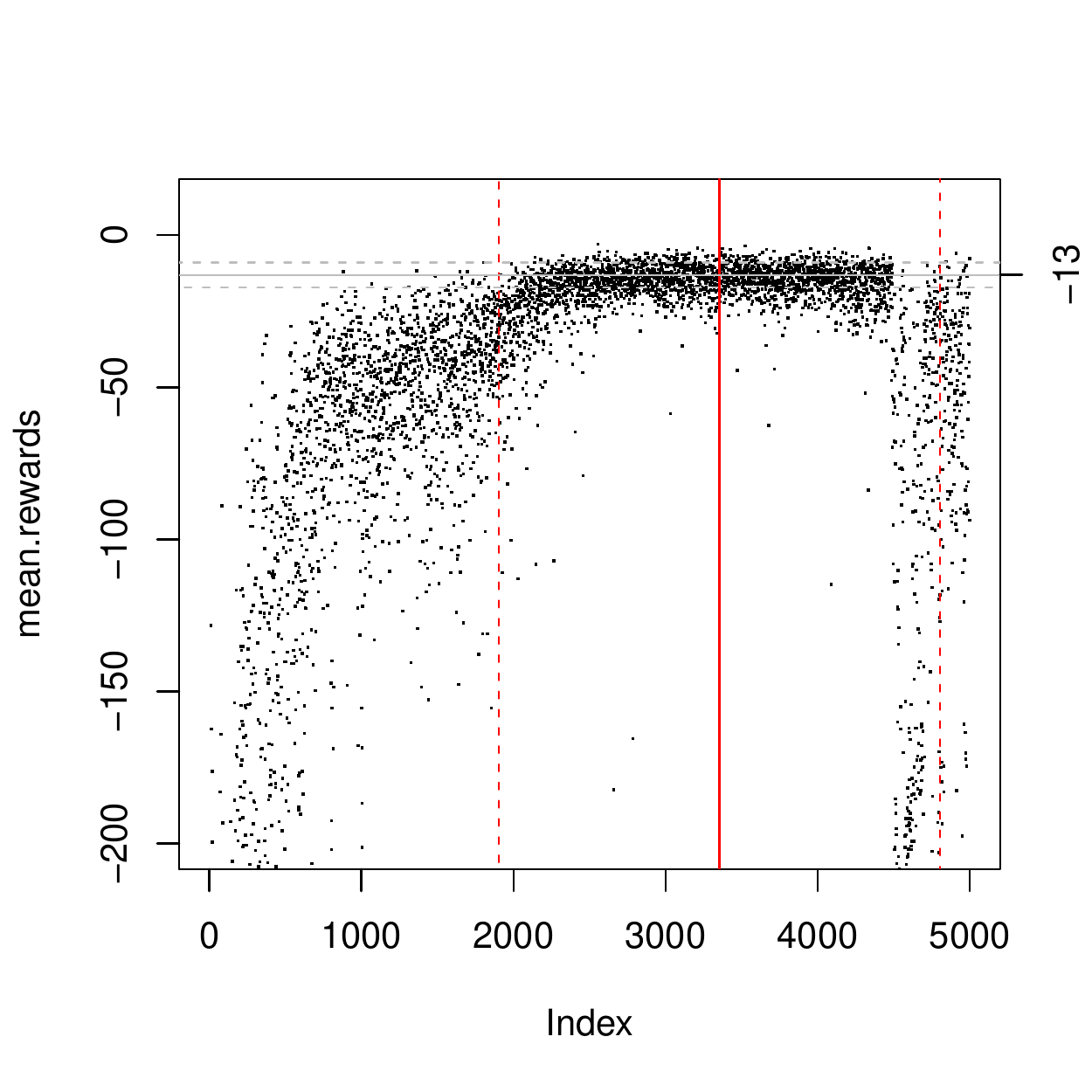}
		\caption{LSTM, partially observable grid world, Advantage.}
		\label{fig:lstmpo}
	\end{subfigure}
	~
	\begin{subfigure}[h]{0.30\textwidth}
		\centering
		\includegraphics[width=\textwidth]{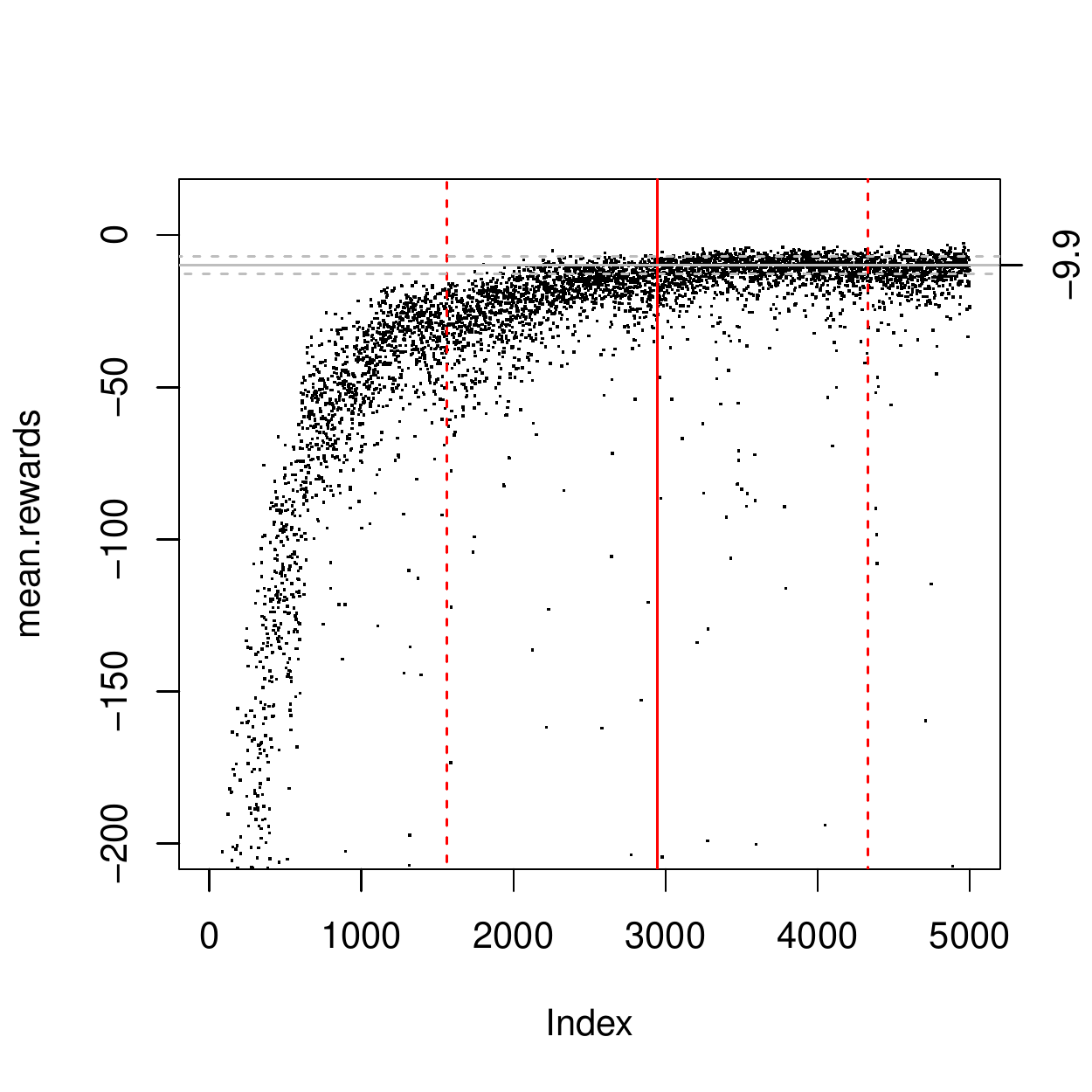}
		\caption{GRU, partially observable grid world, Advantage.}
		\label{fig:grupo}
	\end{subfigure}
	~
	\begin{subfigure}[h]{0.30\textwidth}
		\centering
		\includegraphics[width=\textwidth]{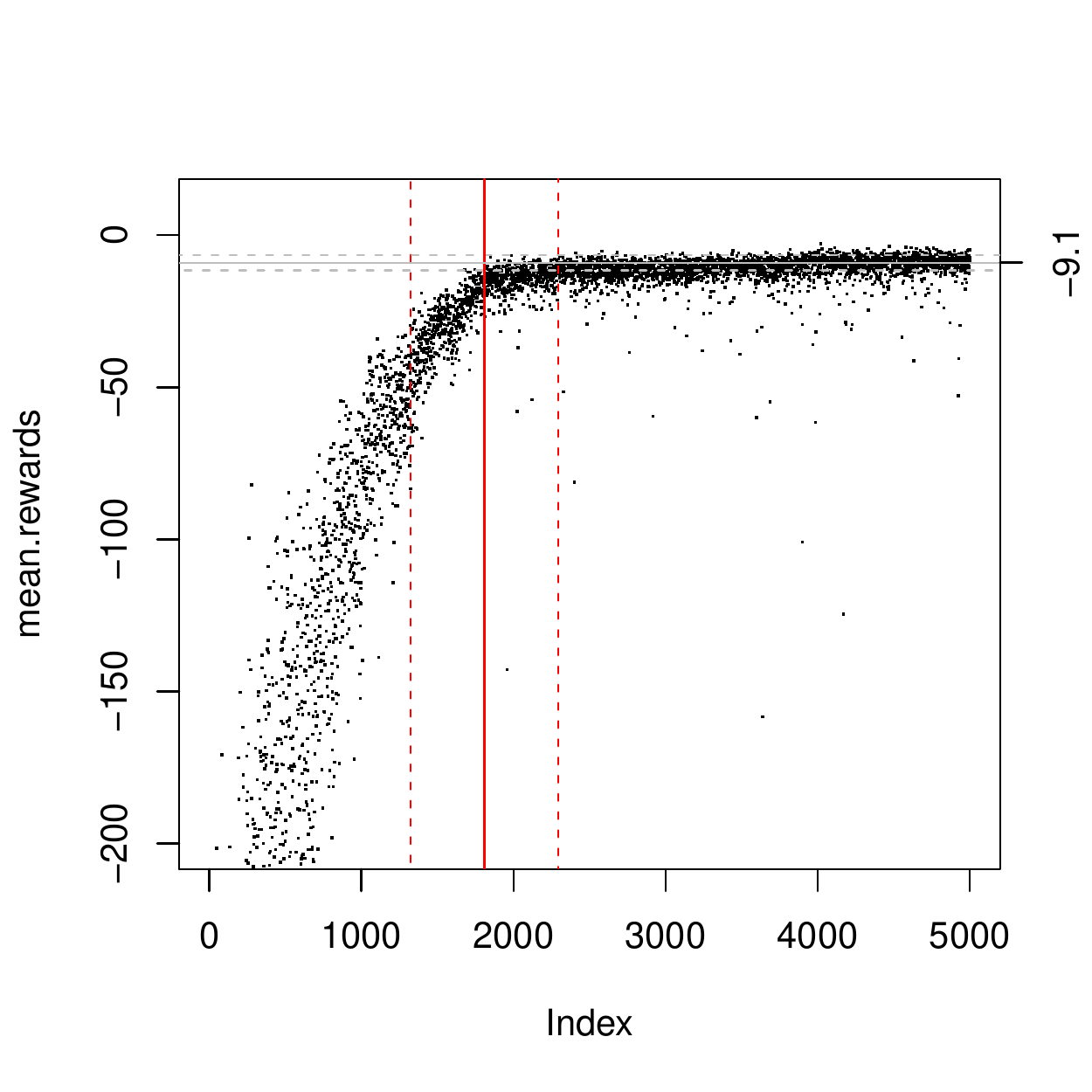}
		\caption{GRU, partially observable grid world, Q-Learning.}
		\label{fig:grupoqlearning}
	\end{subfigure}
	\caption{Average reward over \runs runs for each episode. In the partially observable grid world, Q-Learning allows higher rewards using GRU and LSTM (GRU shown). Using Advantage, GRU and LSTM need more episodes before learning.}
	\label{fig:advantagevsqlearning}
\end{figure}

Advantage Learning leads to smaller learning times and standard deviations than Q-Learning in all worlds except the partially observable grid world. Figure \ref{fig:advantagevsqlearning} shows the behavior of the Q and Advantage learning algorithms in the partially observable grid world. Q-Learning allows faster convergence with a smaller standard deviation.

When using a fixed initial position, GRU learns faster than any other network. The difference of learning speed between GRU and LSTM is statistically significant for (gw, Advantage), (gw, Q-Learning), (po, Q-Learning) and (ac, Q-Learning) (p-values of 0.003, 0.008, 0.0003 and 0.0003, respectively), but not for (po, Advantage) and (ac, Advantage) (p-values of 0.118 and 0.140, respectively).

When using a random initial position, GRU is the only model allowing learning in all the environments when Advantage Learning is used. LSTM and GRU give comparable results in the partially observable worlds, with no statistically significant difference between them.

Agents using MUT1 as a function approximator nearly always manage to learn a good enough policy in partially observable worlds, but they need a large number of episodes to do so. However, plain perceptron-based agents don't manage at all to learn a policy in these worlds\footnote{Except in the partially observable grid world using Q-Learning and random initial positions, where the agent learns to go left, then randomly go up and down until the goal is reached by chance.}, which shows that MUT1 allows better learning in partially observable worlds than a simple non-recurrent neural network.

Table \ref{fig:learningperformance} shows the learning performance of the different neural networks, with the highest values highlighted in bold. Results are displayed in a mean/stddev format.

The feed-forward neural network always achieves the best scores in the grid world, followed by GRU, then LSTM, and finally MUT1. GRU always outperforms the other network architectures in the partially observable worlds. In these worlds, GRU is statistically significantly better than LSTM in all cases (p-value less than 0.0001) except in (random initial, ac, Q-Learning) (p-value of 0.682).

\section{Conclusion}
\label{sec:conc}

LSTM, GRU and MUT1 have been compared on simple reinforcement learning problems. It has been shown that agents using LSTM and GRU for approximating Q or Advantage values perform significantly better than the ones using MUT1, obtaining higher rewards and learning faster.

GRU and LSTM provide comparable performance, with GRU often being significantly better than LSTM. LSTM is never significantly better than GRU. When considering the rewards received by the agents once they have learned, and not the time required for learning, GRU always achieves better results than LSTM.

This shows that using GRU instead of LSTM should be considered when tackling reinforcement problems. Moreover, on the machine used for the experiments, the simpler GRU cell (compared to LSTM) allowed the GRU-based agents to complete their \numepisodes episodes approximately two times faster than LSTM-based agents.

\nocite{Tange2011a}

\bibliographystyle{abbrv}
\bibliography{mybibfile}

\end{document}